\documentclass[review]{elsarticle}

\usepackage{amsmath,amsfonts,amssymb,bm}
\usepackage{graphicx}
\usepackage{booktabs}
\usepackage{color}
\usepackage{lineno}
\usepackage{caption}
\usepackage{subcaption}
\usepackage{ulem}
\usepackage{algpseudocode}
\usepackage{tabularx}
\usepackage{colortbl}
\usepackage{hhline}
\usepackage{physics}
\usepackage{xcolor}
\usepackage[ruled,vlined]{algorithm2e}

\usepackage{lineno,hyperref}
\modulolinenumbers[5]

\journal{Journal of \LaTeX\ Templates}

\begin{document}

\begin{frontmatter}

\title{Data driven Dirichlet sampling on manifolds}

\author{Luan S Prado\fnref{myfootnote} and Thiago G Ritto\fnref{myfootnote}}
\address{Department of Mechanical Engineering, Universidade Federal do Rio de Janeiro}

\begin{abstract}
This article presents a novel method to sampling on manifolds based on the Dirichlet distribution. The proposed strategy allows to completely respect the underlying manifold around which data is observed, and  to do massive samplings with low computational effort. This can be very helpful, for instance, in neural networks training process, as well as in uncertainty analysis and stochastic optimization. Due to its simplicity and efficiency, we believe that the new method has great potential. Three manifolds (two dimensional ring, Mobius strip and spider geometry) are considered to test the proposed methodology, and then it is employed to an engineering application, related to gas seal coefficients.
\end{abstract}


\end{frontmatter}

\section{Introduction}

Machine Learning encompasses several methods and algorithms, from a simple linear regression \cite{Weisberg1985,Hastie2009} to intricate neural networks structures \cite{Haykin2009,Nielsen2015,Ivan2017}. In the past few years, artificial neural networks (ANNs) showed versatility, being able to perform many tasks, such as facial recognition \cite{Lawrence1997} and autonomous vehicle control \cite{Kocic2019}. However, to perform such tasks the training set must be big, since the error surface in neural networks, with large degrees of freedom, tends to be highly non-convex and non-smooth \cite{Khamaru2018}.

In order to circumvent this difficulty, there are some strategies that can be pursued to augment data. This might be helpful in ANNs training, uncertainty quantification, and stochastic optimization. The present work is particularly interested in manifold learning \cite{Ma2012}. Manifold learning shines when the dataset size is small, since it can unravel the intrinsic structure of data \cite{Soize2016,Ghanem2018}. One recent procedure developed to perform probabilistic sampling on manifolds (PSoM) considers multidimensional kernel-density estimation, diffusion maps, and the It\^o stochastic differential equation \cite{Soize2016}. Another strategy explicitly estimates the manifold by the density ridge, and generates new data by bootstrapping \cite{Zhang2020}.

The main contribution of the present article is to present a novel data driven sampling based on the Dirichlet distribution. The proposed Dirichlet sampling on manifolds (DSoM) is straightforward, simple to implement, and efficient to reproduce samples of data around a manifold. Some reference data are needed, which can be obtained directly from a physical system, or be generated with high fidelity models. Then, some points of the original data are randomly selected, the parameters of the Dirichlet distribution are obtained, new data points are generated, and a convex combination is considered to make sure that each point sampled lies around the manifold delineated by the original data.

The proposed strategy is employed to three manifolds: two dimensional ring, Mobius strip and spider geometry. The results are quite satisfactory. Further on, the new method is employed to an engineering application, where a physics-based model is used to compute the pressure distribution inside a gas seal \cite{SanAndres2010}; then the eight seal coefficients (stiffness and damping) are obtained from the pressure field. The simulation points generated by a stochastic physics-based model are used to train an ANN (regression problem). Since simulations of the physics-based system are time consuming, the DSoM is employed to augment the data for the ANN training. The results are again reasonably good, with the added sampling improving the ANN performance.

The organization of this article goes as follows. Section 2 presents the new method, where the context, the main ideas, and the algorithm are discussed. The numerical results are shown in section 3, where three simple manifolds are analyzed, and the engineering application is presented. Finally, the concluding remarks are make in the last section.

\section{Data driven Dirichlet sampling on manifolds -- DSoM}

\subsection{Manifold learning}

Before describing the methodology, it is worth to briefly discuss what is Manifold Learning and which areas it encompasses. Manifold Learning is a multidisciplinary area that involves General Topology \cite{Elon1970,James1999}, Differential Geometry \cite{Spivak1965,Pressley2010} and Statistics \cite{Casella2002,DeGroot2002}.

The main focus of Manifold Learning is the information extraction of manifolds, which are a generalization of curves and surfaces in two, three or higher dimensions spaces. To properly develop the intuition behind the manifold, imagine an ant crawling on a guitar body. For the ant, due to its tiny size, the guitar seems flat and featureless, although its  shape is curved. A manifold is a topological space that locally looks flat and featureless and behaves like an Euclidean Space; however, different from Euclidean Spaces, topological spaces might not have the concept of distance.

In order to clarify what a locally Euclidean space is, some definitions are necessary. A topological space $X$ is said to be locally Euclidean if there exists an integer $d \geq 0$ such that around every point in $X$, there is a local neighborhood which is homeomorphic, i.e. there is a invertible continuous map $g: X \to Y$, to an open subset in an Euclidean space $\mathbb{R}^d$ \cite{Ma2012}.

To extract features from a Manifold one can apply  Isomap \cite{Tenenbaum2000},  Local Linear Embedding (LLE) \cite{Rowies2000}, Laplacian Eigenmaps \cite{Belkin2002}, Diffusion Maps \cite{Nadler2005}, Hessian Eigenmaps \cite{Donoho2003}, and Nonlinear PCA \cite{Scholkopf1998}. 

\subsection{Sampling on manifolds}

{Recently, Soize and Ghanem \cite{Soize2016} developed a method to perform probabilistic sampling on manifolds (PSoM). This strategy is used to generate stochastic samples that follow the probability distribution underlined by a cloud of points concentrated around a manifold, and is based on multidimensional kernel-density estimation, diffusion-maps, and It\^o stochastic differential equation. To avoid MCMC sampling, Zhang and Ghanem \cite{Zhang2020} developed a different strategy that explicitly estimates the manifold by the density ridge, and generates new data by bootstrapping.}

In the present paper we develop a simple and efficient strategy with the same purpose. The two important ingredients in the proposed data driven Dirichlet Sampling on Manifolds (DSoM) are (1) the Dirichlet distribution and (2) convex combinations.

\begin{figure}[!htb]
\begin{center}
\includegraphics[scale=0.5]{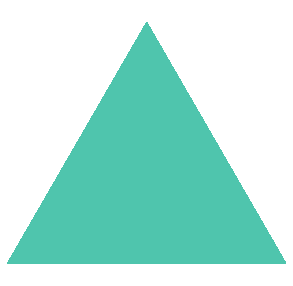}(a)
\includegraphics[scale=0.5]{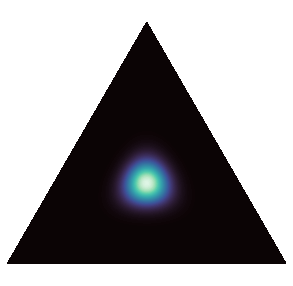}(b)
\includegraphics[scale=0.5]{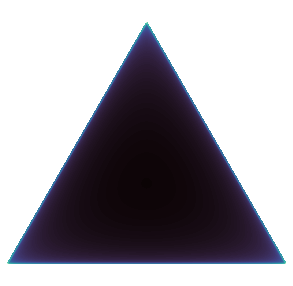}(c)
\includegraphics[scale=0.5]{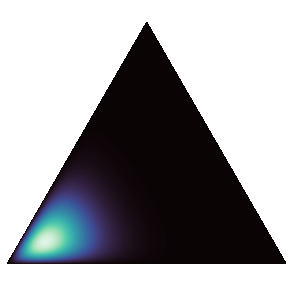}(d)
\end{center}
\caption{Dirichlet distribution with (a) $\alpha = (1, 1, 1)$ (samples are uniformly distributed), (b) $\alpha =(20, 20, 20)$ (as $\alpha$'s increase, the samples get concentrated in the simplex  center), (c)  $\alpha =(0.9, 0.9, 0.9)$ (as $\alpha$'s decrease, the samples get concentrated in the simplex borders), and (d) $\alpha = (10, 2, 2)$ (samples are attracted to a simplex vertex).\label{5}}
\end{figure}

The Dirichlet distribution \cite{MacKay2003} is widely used in other fields, such as text classification \cite{Blei2003}. It is also used in Bayesian Bootstrap \cite{Rubin1981}. This distribution is described by \cite{Kotz2000}:

\begin{equation}
f(\mathbf{x}) =\frac{1}{B(\boldsymbol{\alpha})}\prod_{i=1}^K x_i^{\alpha_i-1}\,,
\end{equation}

\noindent where $\mathbf{x} =(x_1,...,x_K) \in \mathbb{R}^{K}$, with parameters $\boldsymbol{\alpha}=(\alpha_1,...,\alpha_K)$, and the beta function $B$ defined by:

\begin{equation}
{B(\boldsymbol{\alpha})=\frac{\prod_{i=1}^{K} \Gamma(\alpha_i)}{\Gamma(\sum_{i=1}^{K} \alpha_i)}}\,,
\end{equation}

\noindent in which $\Gamma$ is the gamma function, and $\mathbf{x}$ belongs to a $K-1$ simplex, i.e., $\sum_{i=1}^K {x}_{i}=1$ and ${x}_{i}\ge 0$, for $i=1,...,K$;  exactly the same properties needed to guarantee convex combinations.  Samples from the Dirichlet distribution can be concentrated in specific regions of a simplex, depending on the parameter $\boldsymbol{\alpha}$. Some examples are given in Fig.
\ref{5}.

\begin{figure}[!htb]
\centering
\includegraphics[scale=0.12]{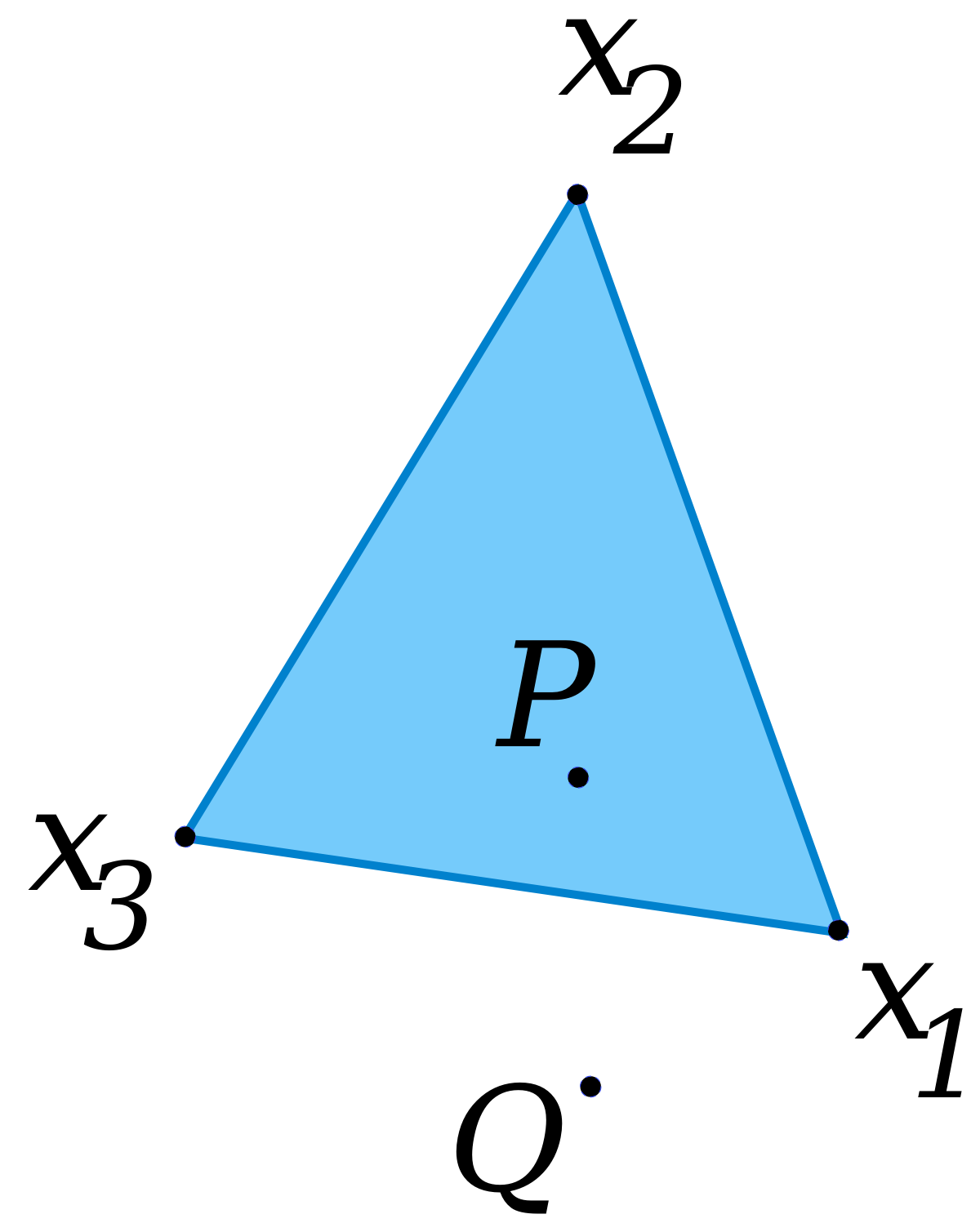}
\caption{$P$ is a convex combination of $x_1, x_2, x_3$,  while the triangle $x_1 x_2 x_3$ is the set of all possible convex combinations of $x_1, x_2, x_3$,  being then its convex hull \cite{wiki}. Note that $Q$ is not a convex combination of $x_1, x_2, x_3$, since it cannot be represented as a convex combination of aforementioned points.}
\label{1}
\end{figure}

Convex combination is a linear combination of the following type: $y = \sum_i \alpha_i x_i,  \alpha_i \geq 0, \sum \alpha_i = 1$. The set of all convex combinations define the convex hull of the $x_i$ points, see Fig. \ref{1}. 

To start the process, we need to have access to a certain amount of data ($m$ samples of a random vector of size $n$) that is organized in a matrix $\mathcal{Y}_{data}\in\mathbb{R}^{m\times n}$, where each line corresponds to one observation:

\begin{equation}
    \mathcal{Y}_{data}=\left(\begin{array}{cccc}
        \mathcal{Y}_{11} & ...& \mathcal{Y}_{1n}\\
        \mathcal{Y}_{21} & ...& \mathcal{Y}_{2n}\\
         & ... & \\
        \mathcal{Y}_{m1} & ...& \mathcal{Y}_{mn}\
    \end{array}\right)=
    \left(\begin{array}{ccc}
        \mathcal{Y}_{(1,:)} \\
        \mathcal{Y}_{(2,:)} \\
         ... \\
        \mathcal{Y}_{(m,:)} \\
    \end{array}\right)\,.
\end{equation}

From the unknown data distribution, we want to generate $n_s$ simulated samples:

\begin{equation}
    \mathbf{Y}_{sim}=
    \left(\begin{array}{ccc}
        \mathbf{Y}_{(1,:)} \\
        \mathbf{Y}_{(2,:)} \\
         ... \\
        \mathbf{Y}_{(n_s,:)} \\
    \end{array}\right)\,,
\end{equation}

\noindent where $\mathbf{Y}\in\mathbb{R}^{n_s\times n}$. With the original data $\mathcal{Y}_{data}$, obtained from a specific application (that can be normalized if necessary), the steps of the proposed methodology are the following.

First we choose randomly (Uniform distribution) $K$ points from the data $\mathcal{Y}_{(i,:)}$ ($i=1,...,K)$. Then, we choose randomly (Uniform distribution) one pivot point $\mathcal{Y}_p$ from the $K$ points. Afterwards, the parameter of the Dirichlet distribution are computed: ${\alpha}_i =\exp\left(-\gamma ||\mathcal{Y}_{(i,:)}-\mathcal{Y}_p||^2\right)$, where $\gamma$ is a tradeoff parameter. Note that, at the pivot, ${\alpha}_i$ equals to one, and it gets smaller when the distance from the pivot increases.

We need to define a threshold $t_{hr}$ and set $\alpha_i=0$ if $\alpha_i<t_{hr}$. This reinforces sampling around the original data, avoiding sampling in void spaces. After that we use the parameters ${\alpha_i}$ to sample $\mathbf{X}=(X_1,...,X_K)\in\mathbb{R}^K$ from the Dirichlet distribution, i.e. $\sum_{i=1}^bX_i=1$. Finally. the $j$-th sample is generated by means of a convex combination \cite{Tyrrel1970} of the $K$ data points: $\mathbf{Y}_{(j,:)}=X_1\mathcal{Y}_{(1.:)}+...+X_K\mathcal{Y}_{(K.:)}$. Note that the Dirichlet sample serves as weights to each one of the $K$ points. To generate a new sample, the process is repeated with the random selection of other $K$ points from the data $\mathcal{Y}_{(i,:)}$ ($i=1,...,K)$.


Thus, we need to tune only three parameters: $K$, which is the number of data points used in the process of generating each simulated sample; $\gamma$, that defines the shape of the exponential curve; and $t_{hr}$, which is the threshold that will define zero  weight for points far away from the pivot. We also need to define the number of samples $n_s$ and a metric to compute the norm $||\cdot||$; for instance, the Euclidean norm or the Mahalanobis distance.

Finally, it should be noted that each sample requires the computation of an Algebraic system $\mathbf{Y}_{(j,:)} = A^T\mathbf{X}$, in which $\mathbf{Y}_{(j,:)}$ is the sampled point while $A^T\mathbf{X}$ is the product of the dataset points by its weights, sampled from a Dirichlet distribution ($\mathbf{Y}_{(j,:)}\in \mathbb{R}^{ n}$, $\mathbf{X}\in \mathbb{R}^{K}$, $A\in \mathbb{R}^{K\times n}$). In order to avoid cumbersome computations, few samples are used instead of the whole dataset, and the computational complexity is $\mathcal{O}(Kn)$, where $K$ is prescribed by the user and $n$ is the size of the random vector.

\subsection{DSoM algorithm and convergence}

The DSoM algorithm is given below (Algorithm \ref{algorithm}).

\begin{algorithm}[!htb]
\SetAlgoLined
 \For{j = 1 to $n_s$}{
  $A = sample(dataset,K)$
  
  $\mathcal{Y}_p = sample(A,1)$
  
    \For{i = 1 to K}{
        $\alpha[i] = exp(-\gamma||\mathcal{Y}[i]-\mathcal{Y}_p||)$
        
        \eIf{$\alpha[i]<threshold$}
        
        {
        $\alpha[i] = 1e-7$}
        {
        
        $\alpha[i] = \alpha[i]$
	     }
	     
	    {
           }
   }
  $\mathbf{X} = Dirichlet(\alpha_1,...,\alpha_K)$
  
  $\mathbf{Y}[j] = A^T\mathbf{X}$
  }  
  \caption{DSoM(dataset,$n_s$,$K$,$threshold$)}
 \label{algorithm}
\end{algorithm}

To verify if the proposed method is converging to the underlined distribution of the points that lie around a manifold, we check the convergence in mean:

\begin{equation}
    conv_1(n_s) =\displaystyle\frac{\frac{1}{n_s}\sum_{j=1}^{n_s}\left|\left| \mathbf{Y}_{(j,:)} - \left(\frac{1}{m}\sum_{i=1}^{m} {\mathcal{Y}_{data}}_{(i,:)}\right) \right|\right|}{\left|\left| \frac{1}{m}\sum_{i=1}^{m} {\mathcal{Y}_{data}}_{(i,:)} \right|\right|}\,.\label{eq_conv1}
\end{equation}

This equation is considering the mean of the original data point as a reference, and it observes the convergence of the mean value of the Euclidian norm of the random vector. This convergence is not sufficient because we also want to assure the convergence related to the correlation among the random variables. Hence, we also consider the convergence of the correlation matrix, in terms of the Frobenious norm,

\begin{equation}
    conv_2(n_s) =\displaystyle\frac{\left|\left|\frac{1}{n_s-1} \mathbf{Y}^T \mathbf{Y} - \frac{1}{m-1}\sum_{i=1}^{m} \mathcal{Y}_{data}^T\mathcal{Y}_{data} \right|\right|_{F}}{\left|\left| \frac{1}{m-1}\sum_{i=1}^{m} \mathcal{Y}_{data}^T\mathcal{Y}_{data} \right|\right|_{F}}\,.\label{eq_conv2}
\end{equation}

Indeed, since we are dealing with manifolds, one must be careful with these convergence metrics, and also observe the simulated points.

\section{Numerical Results}

\subsection{Three simple manifolds}

\begin{figure}[!htb]
\centering
\includegraphics[scale=0.3]{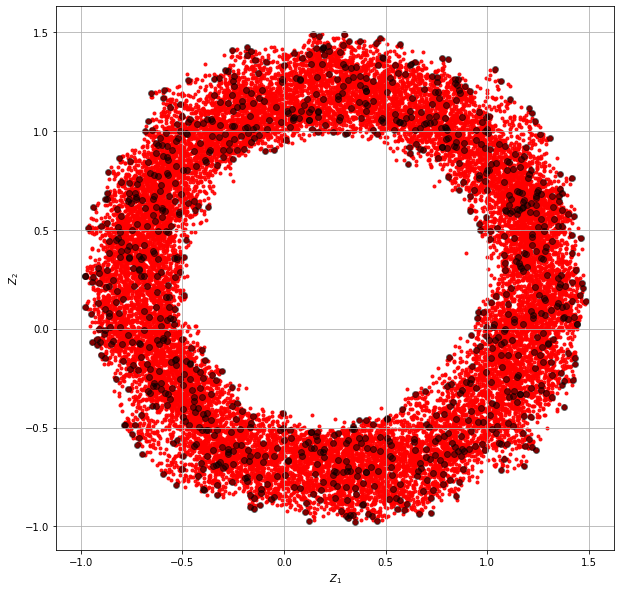}
\caption{Two dimensional ring. 50,000 sampled data (red dots) from 1,000 original database (black dots). Parameters: $K=5$ $\gamma=0.8$, $t_{sh}=0.8$.}
\label{9}
\end{figure}

Before applying the sampling strategy to the engineering application, a verification of the method must be done. Figure \ref{9} shows the results for a two dimensional ring. The original data was generated by

\begin{equation}
\begin{array}{l}
Z_1 = \cos(\theta)+U\,,\\
Z_2=  \sin(\theta)+U\,,\\
\end{array}
\end{equation}

\noindent with $\theta \in[0,2\pi]$, and $U$ is a Uniform random variable with support $[0,0.5]$. The Figure \ref{9} shows $m=1,000$ original data points (black dots), and $n_s=50,000$ new data (red dots) sampled with parameters $K=10$, $\gamma=0.8$ and $t_{sh}=0.8$. It is noticed that the simulated data yielded by DSoM respect the original cloud of points observed around the manifold.

\begin{figure}[!htb]
    \centering
    \includegraphics[scale=.39]{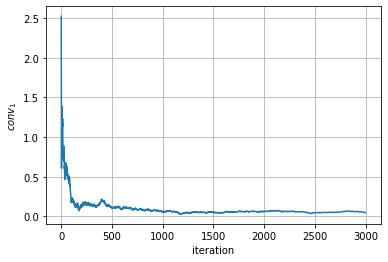}(a)
    \includegraphics[scale=.64]{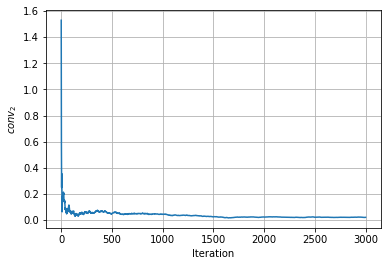}(b)
    \caption{Convergence curves: (a) $conv_1(n_s)$, $L_2$-norm of the mean, and (b) $conv_2(n_s)$, Frobenious-norm of the correlation matrix.}
    \label{conv1}
\end{figure}

\begin{figure}[!htb]
\centering
\includegraphics[scale=0.65]{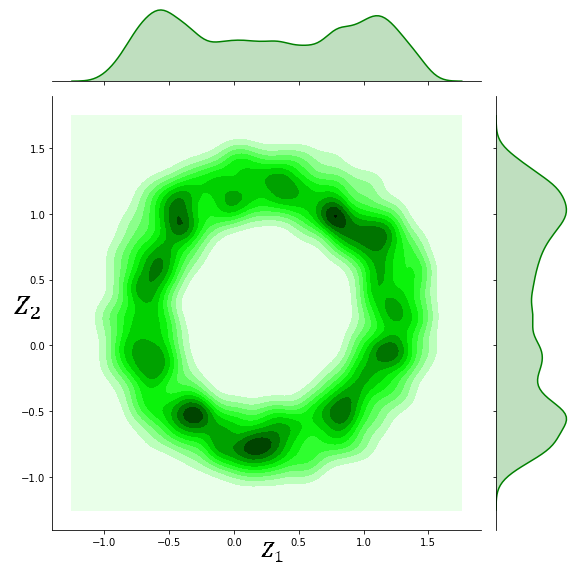}
\caption{Kernel Density Estimation of the red dots shown in Fig. \ref{9}.}
\label{10}
\end{figure}

Figure \ref{conv1} shows the convergence of the mean and the correlation matrix; Eqs. (\ref{eq_conv1}) and (\ref{eq_conv2}). It can be seeing that the convergence is quite reasonable. In addition, Fig. \ref{10} shows a heat map, obtained by kernel density estimation, where the marginal distribution (clearly non-Gaussian) are plotted.

\begin{figure}[!htb]
\centering
\includegraphics[scale=0.64]{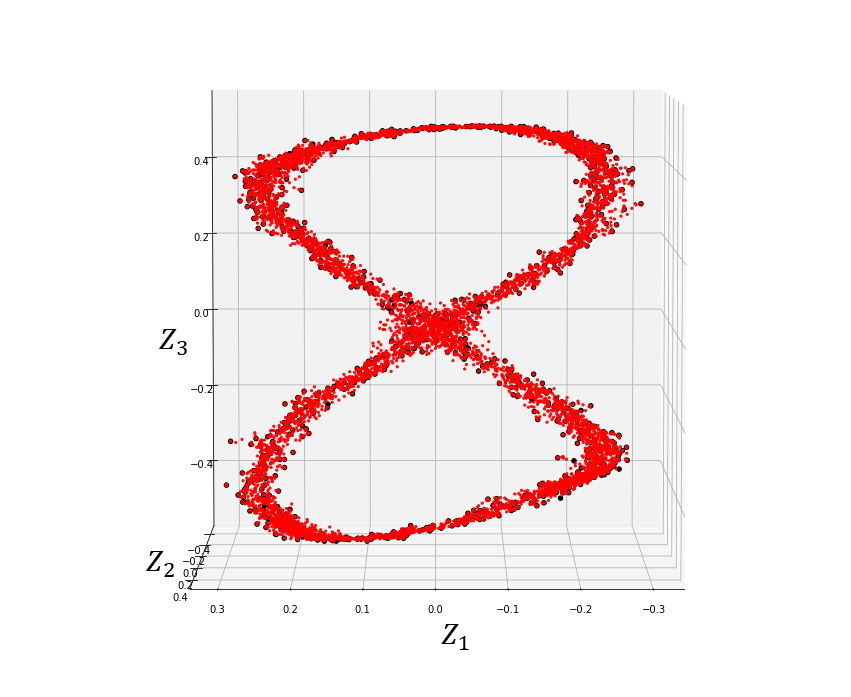}
\caption{Mobius strip. 10,000 sampled data (red dots) from 1,000 original data (black dots). Parameters: $K=20$ $\gamma=5$, $t_{sh}=0.9$.}
\label{fig10}
\end{figure}

To avail the proposed sampling strategy functionality, two more manifolds are considered. The first is a Mobius Strip, parametrized by 

\begin{equation}
\begin{array}{l}
Z_1 = 0.5 \sin(0.5t)\cos(0.5t)+\epsilon\,,\\
Z_2=  0.5 \cos(0.5t) \cos(0.5t)+\epsilon\,,\\
Z_3 = 0.5 \cos(0.5)+\epsilon\,,
\end{array}
\end{equation}

\noindent with $t\in[0,8\pi]$, and $\epsilon$ is a Gaussian random variable with zero mean and standard deviation equals to 0.05. Figure \ref{fig10} shows $m=1,000$ original data points (black dots), and $n_s=10,000$ new data (red dots) sampled with parameters $K=20$, $\gamma=5$ and $t_{sh}=0.9$. Again, the simulated data generated by DSoM respect the original cloud of points observed around this manifold.

The last manifold tested is the spider geometry obtained from a ply file \cite{site}. The results are shown in Fig. \ref{14}, where the DSoM was applied with parameters $K=300$ $\gamma=0.85$, $t_{sh}=0.7$. Note that, spite of the challenging geometry, the method still works well.

\begin{figure}[!htb]
\centering
\includegraphics[scale=0.64]{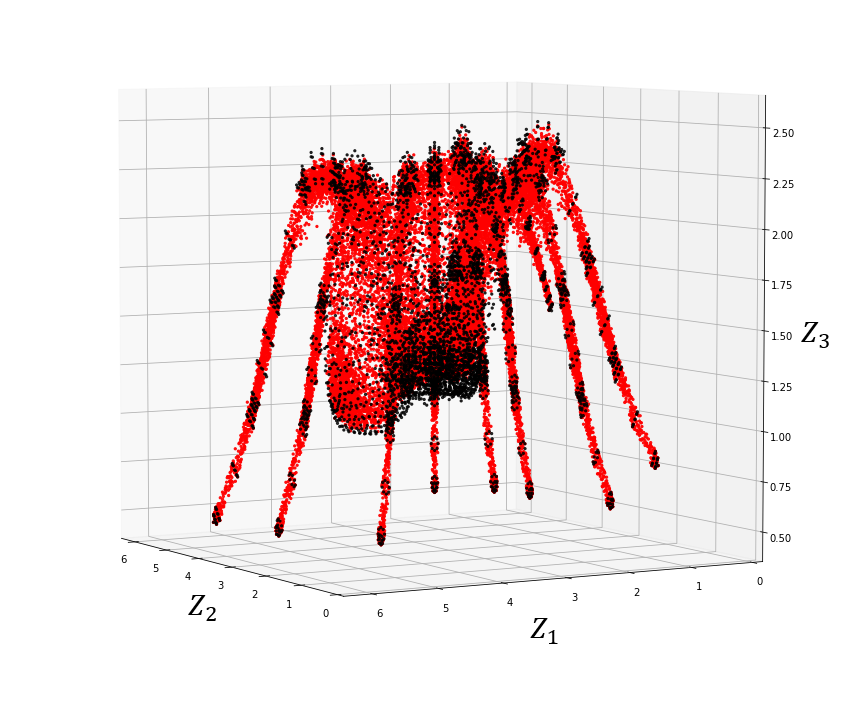}
\caption{Spider geometry. 10,000 sampled data (red dots) from 5,000 original data (black dots). Parameters: $K=300$ $\gamma=0.85$, $t_{sh}=0.7$.}
\end{figure}

In the next section, we apply the methodology to an engineering problem, related to the coefficients of a gas seal, where eight parameters are considered, i.e., the dimension of the random vector is $n=8$.

\subsection{Gas seal coefficients}

Before applying the DSoM, we need to explain the steps of the analysis performed in this section. We are interested in computing the eight seal coefficients of a centrifugal compressor. For this purpose, the Reynolds equation, for a compressible fluid, is used \cite{Gross1962,Pan1980,Hamrock1994,Kleynhans1996,Faria2000,SanAndres2010}:

\begin{equation}
\frac{ \partial}{\partial \bar{x}} {\left( PH^3 \frac{ \partial P}{\partial \bar{x}}\right)}  + \frac{ \partial}{ \partial \bar{z}} {\left( PH^3 \frac{  \partial P}{ \partial \bar{z}} \right)} = \lambda {\frac{\partial (PH)}{ \partial \bar{x}}}+\sigma {\frac{\partial (PH)}{\partial \tau}}\,,
\end{equation}

\noindent with $P = p/p_a$, $H = h/h^{*}$, $\tau = \omega t$,  $\bar{x} = x /L^{*}$, $\bar{z} = z /L^{*}$, $\lambda = \frac{6\mu U L^*}{p_a (h^*)^2}$, and $\sigma = \frac{12\mu\omega (L^*)^2}{p_a (h^*)^2} $; where $L^{*}$ is a characteristic length of the bearing,  $h^{*}$ is a characteristic film thickness, $p_a$ is the atmospheric pressure, $\omega$ the rotation speed of the shaft. 

The pressured field and the corresponding seal coefficients are computed using the ISOTSEAL (constant-temperature seal code) \cite{Kleynhans1996,Holt2002}. The idea is to use this physics-based model to train an ANN, that will serve as a surrogate model for the system under analysis.

The following training procedure was adopted. A stochastic model is built, considering some parameters of the deterministic model as random variables. This is done to explore the surroundings of the chosen configuration, and to create a more robust ANN, that takes into account uncertainties related to the model parameters. After the initial training with the stochastic model, the DSoM is employed to augment data and leverage the ANN training process.

Twenty model parameters are varied, which means that they are modeled as random variables and serve as input to the ANN: seal radius, number of teeth, tooth pitch, tooth height, radial clearance, gas composition of methane, ethane, propane, isobutan, butane, hydrogen, nitrogen, oxygen, $CO_2$, reservoir temperature, reservoir pressure, sump pressure, inlet tangential velocity ratio, whirl speed, rotational speed. The outputs of the model are the eight seal coefficients (stiffness and damping), namely $k_{xx}$, $k_{xy}$, $k_{yx}$, $k_{yy}$, $c_{xx}$, $c_{xy}$, $c_{yx}$, $c_{yy}$.

Each one of the twenty input variables are modeled with a Uniform distribution, with support that encompass 20\% above and 20\% below the reference value. A normalization process was not carried out; however, depending on the problem and on its dimensionalty, its use is suggested.

The constructed ANN has 20 neurons in the first layer (input parameters), and 8 neurons in the last layer (seal coefficients). There are 2 hidden layers with sixteen neurons each, activated by the ReLU function.

\begin{figure}[!htb]
\centering
\includegraphics[scale=.36]{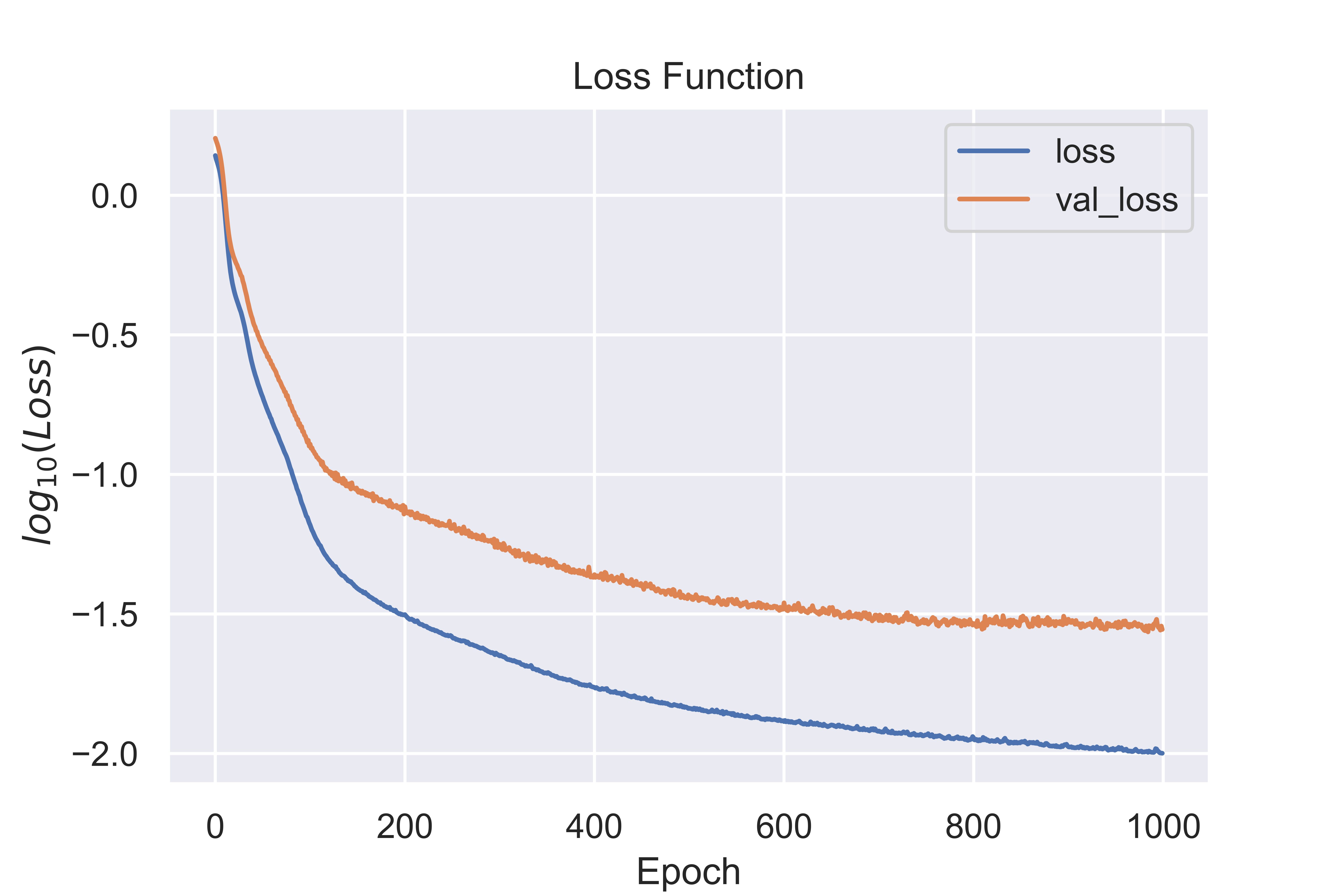}(a)
\includegraphics[scale=.36]{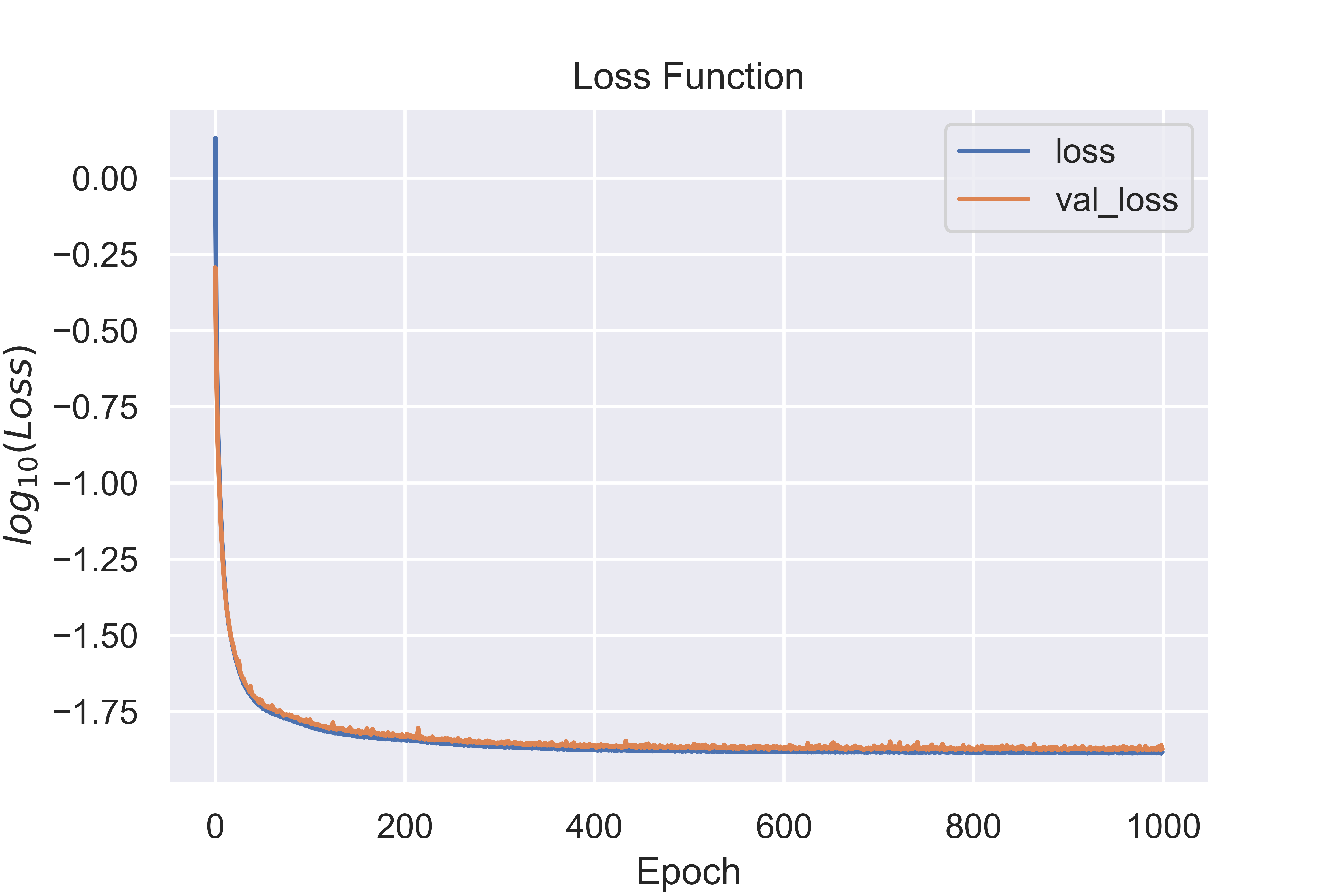}(b)
\caption{Neural network loss function in different scenarios. (a) ANN trained with 4,205 points from the stochastic physics-based model, and (b) ANN trained with 100,000 points, augmenting data using DSoM.}
\label{14}
\end{figure}

The first ANN was trained with 4,205 points sampled from the stochastic physics-based model. The second ANN was trained  with the DSoM, which was sampled from the previous database, where a 100,000 points were generated. In Fig. \ref{14}, the effectiveness of DSoM is exhibited in terms of the loss function. Note that, with the proposed data augmentation procedure, overfit is removed from the ANN. The distance between the loss function in training (blue line) and the loss function in validation (orange line) is close to zero in the second scenario.

\begin{figure}[!htb]
\centering
\includegraphics[scale=0.17]{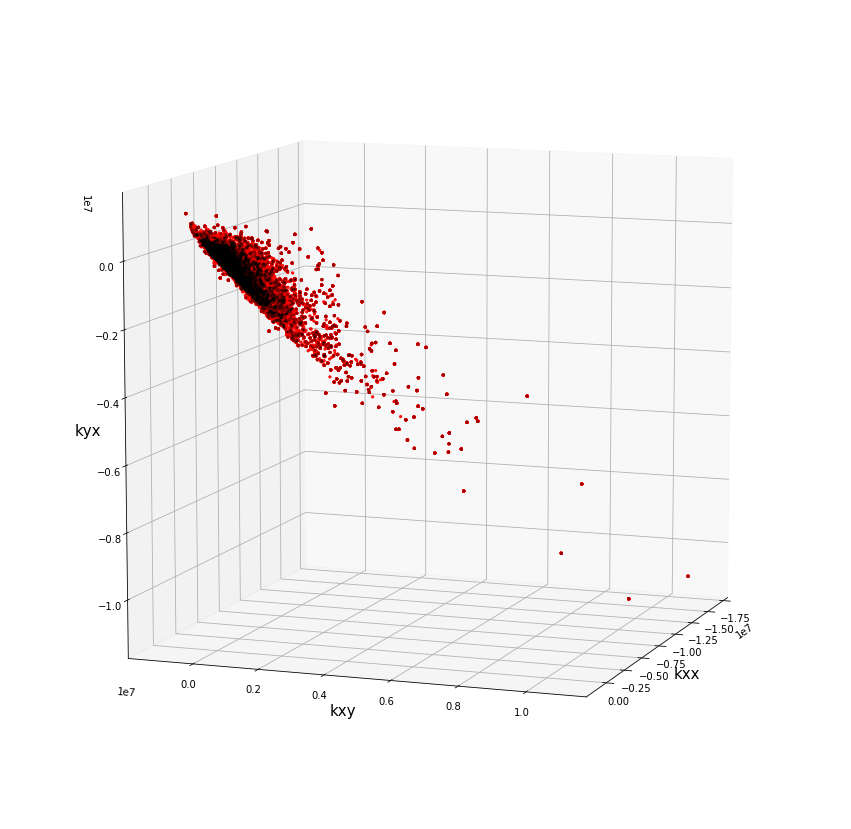}(a)
\includegraphics[scale=0.17]{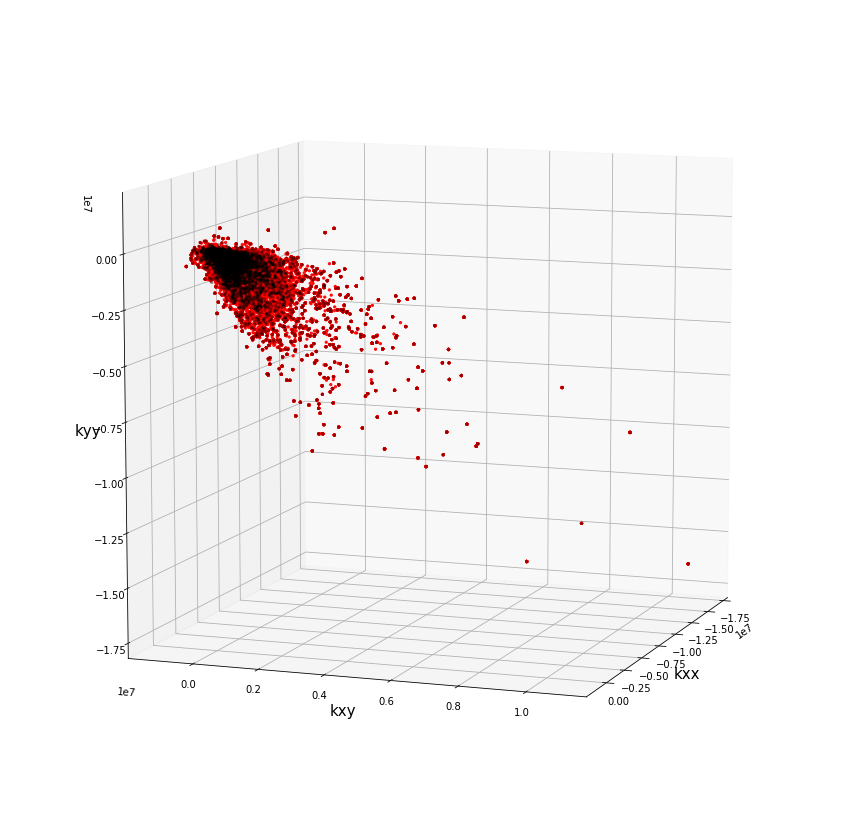}(b)
\includegraphics[scale=0.17]{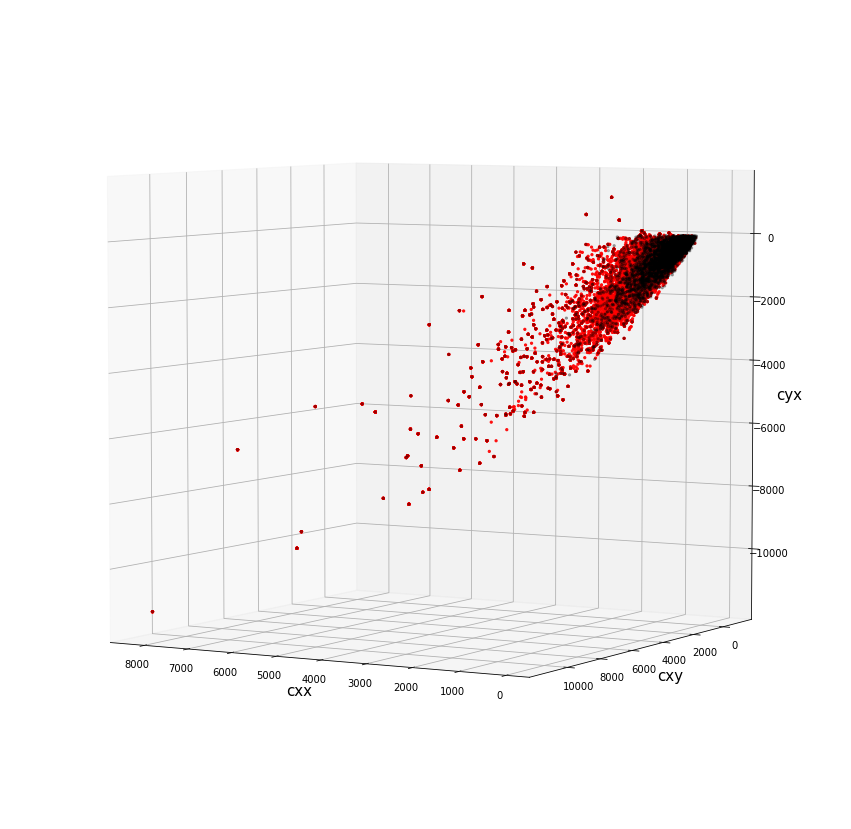}(c)
\includegraphics[scale=0.17]{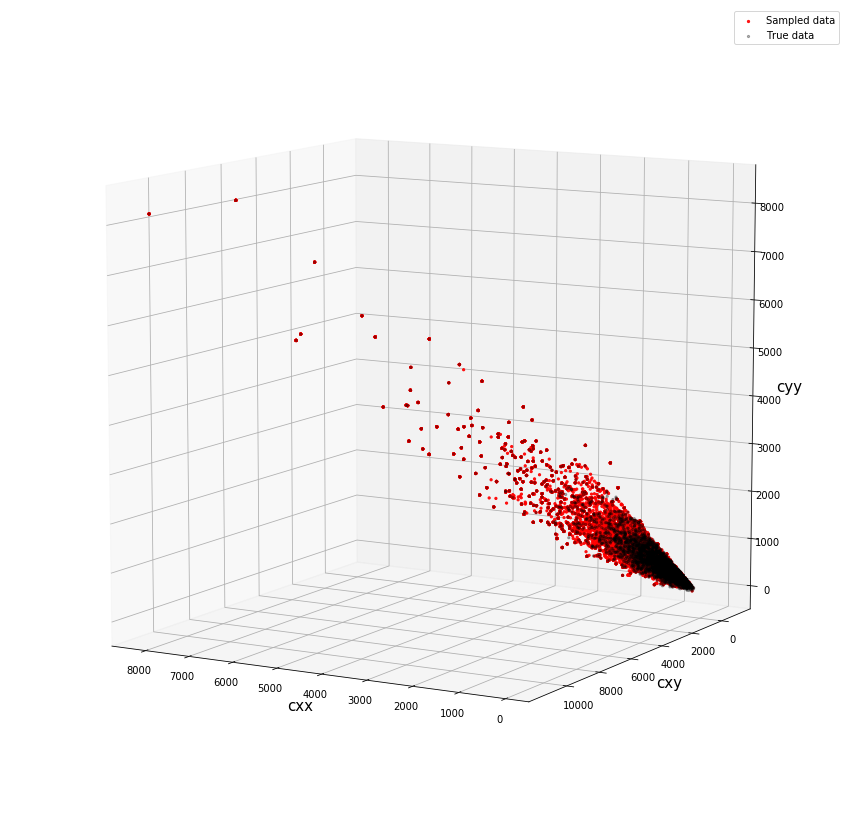}(d)
\caption{stiffness seal coefficients plotted against each other. DSoM parameters: $\gamma = 6e-13$, $t_{hr} = 0.5$}
\label{seal_stiffness}
\end{figure}

Figure \ref{seal_stiffness} shows the samples of the seal coefficients. The black dots were computed using the stochastic physics-based model, and the red dots were obtained taken into account these original data, and employing the DSoM to increase the number of points. Since the dimension of the random vector is greater than three, each graphic in Fig. \ref{seal_stiffness} shows a cloud of points for three different parameters. It can be observed that the manifold structure is respected.

\begin{figure}[!htb]
    \centering
    \includegraphics[scale=.39]{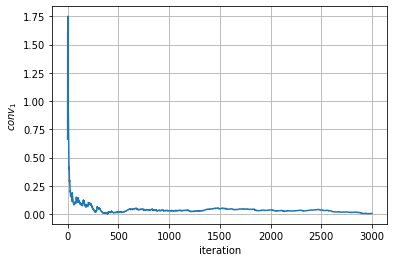}(a)
    \includegraphics[scale=.65]{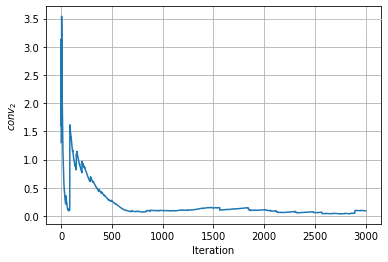}(b)
    \caption{Convergence curves: (a) $conv_1(n_s)$, $L_2$-norm of the mean, and (b) $conv_2(n_s)$, Frobenious-norm of the correlation matrix.}
    \label{conv_seal}
\end{figure}

Figure \ref{conv_seal} shows the convergence of the mean and the correlation matrix; Eqs. (\ref{eq_conv1}) and (\ref{eq_conv2}). It can be seeing that the convergence is also reasonable for this application, with convergence values lower than 5\%.

\section{Concluding remarks}

This paper proposes a new methodology to sample on manifolds using the Dirichilet distribution, which is simple and effective. The Dirichlet sampling on manifolds (DSoM) requires an original sampling that can be obtained from simulation or from experiments. The DSoM generates samples that follow the unknown distribution of the original dataset. This might be helpful in uncertainty quantification, stochastic optimization, and ANN training.

The methodology was successfully applied to three simple manifolds, and an engineering application, related to seal coefficients. The next step is to apply it to different problems to test its versatility. In addition, a formal mathematical proof of the efficiency of the DSoM should be pursued.

\bibstyle{elsarticle-num}
\bibliography{bib_luan}

\begin{thebibliography}{10}
\expandafter\ifx\csname url\endcsname\relax
  \def\url#1{\texttt{#1}}\fi
\expandafter\ifx\csname urlprefix\endcsname\relax\def\urlprefix{URL }\fi
\expandafter\ifx\csname href\endcsname\relax
  \def\href#1#2{#2} \def\path#1{#1}\fi

\bibitem{Weisberg1985}
S.~Weisberg, Applied Linear Regression, 1st Edition, Wiley, 1985.

\bibitem{Hastie2009}
T.~Hastie, Tibishirani, R., Friedman, Elements of Statistical Learning, 1st
  Edition, Springer, 2009.

\bibitem{Haykin2009}
S.~Haykin, Neural Networks and Learning Machines, 1st Edition, Pearson, 2009.

\bibitem{Nielsen2015}
M.~A. Nielsen, Neural networks and deep learning, Determination Press, 2015.

\bibitem{Ivan2017}
I.~N. da~Silva, D.~H. Spatti, R.~A. Flauzino, L.~H.~B. Liboni, S.~F. dos
  Reis~Alves, Artificial Neural Networks: A Practical Course, 1st Edition,
  Springer Publishing Company, Incorporated, 2016.

\bibitem{Lawrence1997}
S.~Lawrence, C.~Giles, A.~C. Tsoi, A.~Back,
  \href{http://ieeexplore.ieee.org/xpls/abs_all.jsp?arnumber=554195&tag=1}{Face
  recognition: a convolutional neural-network approach}, Neural Networks, IEEE
  Transactions on 8~(1) (1997) 98--113.
\newblock \href {https://doi.org/10.1109/72.554195}
  {\path{doi:10.1109/72.554195}}.
\newline\urlprefix\url{http://ieeexplore.ieee.org/xpls/abs_all.jsp?arnumber=554195&tag=1}

\bibitem{Kocic2019}
J.~Kocic, N.~S. Jovicic, V.~Drndarevic,
  \href{http://dblp.uni-trier.de/db/journals/sensors/sensors19.html#KocicJD19}{An
  end-to-end deep neural network for autonomous driving designed for embedded
  automotive platforms.}, Sensors 19~(9) (2019) 2064.
\newline\urlprefix\url{http://dblp.uni-trier.de/db/journals/sensors/sensors19.html#KocicJD19}

\bibitem{Khamaru2018}
K.~Khamaru, M.~Wainwright, Convergence guarantees for a class of non-convex and
  non-smooth optimization problems, in: Proceedings of the 35th International
  Conference on Machine Learning, PMLR, Vol.~80, 2018, pp. 2601--2610.

\bibitem{Ma2012}
Y.~Ma, Y.~Fu, Manifold Learning Theory and Applications, CRC Press, 2012.

\bibitem{Soize2016}
C.~Soize, R.~Ghanem, Data-driven probability concentration and sampling on
  manifold, Journal of Computational Physic 321 (2016) 242--258.

\bibitem{Ghanem2018}
R.~Ghanem, Statistical sampling on manifolds for expensive computational
  models, CDSE Days (2018).

\bibitem{Zhang2020}
R.~Zhang, R.~Ghanem, Normal-bundle bootstrap, arXiv (07 2020).

\bibitem{SanAndres2010}
L.~San~Andr\'es, Modern Lubrication Theory, Gas Film Lubrication,, Texas A \& M
  University Digital Libraries, 2010.

\bibitem{Elon1970}
E.~L. Lima, Elementos de Topologia Geral, 1st Edition, Editora USP, S\~ao
  Paulo, 1970.

\bibitem{James1999}
I.~M. James, History of Topology, Elsevier B. V., Netherlands, 1999.

\bibitem{Spivak1965}
M.~Spivak, Calculus on Manifolds, Benjamin, New York, 1965.

\bibitem{Pressley2010}
A.~Pressley, Elementary Differential Geometry, 2nd Edition, Springer-Verlag,
  New York, 2010.

\bibitem{Casella2002}
G.~Casella, R.~L. Berger, Statistical Inference, Duxbury Press, 2002.

\bibitem{DeGroot2002}
M.~DeGroot, Schervish, Probability and Statistics, 3rd Edition, Addison-Wesley,
  2002.

\bibitem{Tenenbaum2000}
J.~Tenenbaum, V.~Silva, J.~Langford, A global geometric framework for nonlinear
  dimensionality reduction, Science (New York, N.Y.) 290 (2001) 2319--23.
\newblock \href {https://doi.org/10.1126/science.290.5500.2319}
  {\path{doi:10.1126/science.290.5500.2319}}.

\bibitem{Rowies2000}
S.~Roweis, L.~Saul, Nonlinear dimensionality reduction by locally linear
  embedding, Science (New York, N.Y.) 290 (2001) 2323--6.
\newblock \href {https://doi.org/10.1126/science.290.5500.2323}
  {\path{doi:10.1126/science.290.5500.2323}}.

\bibitem{Belkin2002}
M.~Belkin, P.~Niyogi, Laplacian eigenmaps and spectral techniques for embedding
  and clustering, Advances in Neural Information Processing System 14 (04
  2002).

\bibitem{Nadler2005}
B.~Nadler, S.~Lafon, R.~Coifman, I.~Kevrekidis, Diffusion maps, spectral
  clustering and eigenfunctions of fokker-planck operators, Adv Neural Inf
  Process Syst 18 (07 2005).

\bibitem{Donoho2003}
D.~Donoho, C.~Grimes, Hessian eigenmaps: Locally linear embedding techniques
  for high-dimensional data. proc. national academy of science (pnas), 100,
  5591-5596, Proceedings of the National Academy of Sciences of the United
  States of America 100 (2003) 5591--6.
\newblock \href {https://doi.org/10.1073/pnas.1031596100}
  {\path{doi:10.1073/pnas.1031596100}}.

\bibitem{Scholkopf1998}
B.~Schölkopf, A.~Smola, K.-R. Müller, Nonlinear component analysis as a
  kernel eigenvalue problem, Neural Computation 10 (1998) 1299--1319.
\newblock \href {https://doi.org/10.1162/089976698300017467}
  {\path{doi:10.1162/089976698300017467}}.

\bibitem{MacKay2003}
J.~C.~D. MacKay, Information Theory, Inference, and Learning Algorithms,
  Cambridge University Press, 2003.

\bibitem{Blei2003}
D.~Blei, A.~Ng, M.~Jordan, Latent dirichlet allocation, Journal of Machine
  Learning Research 3 (2013) 993.

\bibitem{Rubin1981}
D.~Rubin, The bayesian bootstrap, Ann Statist 9 (1981) 130--134.

\bibitem{Kotz2000}
S.~Kotz, N.~Balakrishnan, N.~L. Johnson, Multivariate Distributions: Reduced
  Models and Applications, Vol.~1, Wiley, 2000.

\bibitem{wiki}
Wikipedia, Convex combinations,
  \url{https://en.wikipedia.org/wiki/Convex_combination} (October 2020).

\bibitem{Tyrrel1970}
R.~R. Tyrrel, Convex Analysis, Princeton University Press, 1970.

\bibitem{site}
J.~Burkardt, Ply files an ascii polygon format,
  https://people.sc.fsu.edu/~jburkardt/data/ply/ply.html (June 2012).

\bibitem{Gross1962}
W.~Gross, Gas Film Lubrication,, John Wiley \& Sons, Inc., 1962.

\bibitem{Pan1980}
C.~Pan, Gas Bearing Tribology: Friction, Lubrication and Wear, A.Z. Szeri,
  Hemisphere Pub. Corp., 1980.

\bibitem{Hamrock1994}
B.~Hamrock, Fundamentals of Fluid Film Lubrication, McGrawHill, Inc., 1994.

\bibitem{Kleynhans1996}
G.~Kleynhans, A two-control-volume bulk-fow rotordynamic analysis for
  smooth-rotor/honeycomb-stator gas annular, Ph.D. thesis, Texas A\& M
  University (1996).

\bibitem{Faria2000}
M.~Faria, L.~San~Andr\'es, On the numerical modeling of high speed hydrodynamic
  gas bearings, ASME Journal of Tribology 122~(1) (2000) 124--130.

\bibitem{Holt2002}
C.~Holt, D.~W. Childs, Theory versus experiment for the rotordynamic impedances
  of two hole-pattern-stator gas annular seals, Journal of Tribology 124~(1)
  (2002) 137--143.

\end{thebibliography}

\end{document}